\pdfoutput=1
\documentclass[11pt]{article}

\usepackage[]{ACL2023}

\usepackage{times}
\usepackage{latexsym}
\usepackage{microtype}
\usepackage{setspace}

\usepackage{hyperref}

\usepackage{graphicx}
\usepackage{soul}
\usepackage{url}

\usepackage{amsmath}
\usepackage{amsthm}
\usepackage{booktabs}
\usepackage[ruled, linesnumbered]{algorithm2e}
\usepackage{array}
\usepackage{color, colortbl}
\usepackage{multirow}
\usepackage{setspace}
\usepackage{autobreak}
\usepackage{amsfonts}
\usepackage{amssymb}
\usepackage{amsthm,amsmath}
\usepackage{mathrsfs}
\usepackage{multirow}
\usepackage{subfigure}
\usepackage{makecell}
\usepackage{pifont}
\usepackage{fontawesome}
\usepackage{bm}

\usepackage[T1]{fontenc}
\usepackage[utf8]{inputenc}

\usepackage{microtype}

\usepackage{inconsolata}

\title{DSRM: Boost Textual Adversarial Training with Distribution Shift Risk Minimization}

\author{
    Songyang Gao$^{1}$, \ \ Shihan Dou$^{1}$, \ \ Yan Liu$^{1}$,\ \  Xiao Wang$^{1}$,\ \ Qi Zhang$^{12}$\thanks{{ }{ }{ }Corresponding author.} ,\\
    \textbf{ Zhongyu Wei$^3$, \ \ Jin Ma$^4$, \ \ Ying Shan$^4$} \\ 
    \normalsize{$^1$  School of Computer Science, Fudan University,\ Shanghai,\ China} \\
    \normalsize{$^2$  Shanghai Key Laboratory of Intelligent Information Processing,\ Shanghai,\ China} \\
    \normalsize{$^3$School of Data Science,\ Fudan University,\ Shanghai,\ China  \ \  $^4$Tencent PCG} \\
    \normalsize{ \{gaosy21, shdou21\}@m.fudan.edu.cn}
}

\begin{document}
\maketitle

\begin{abstract}
Adversarial training is one of the best-performing methods in improving the robustness of deep language models. 
However, robust models come at the cost of high time consumption, as they require multi-step gradient ascents or word substitutions to obtain adversarial samples. In addition, these generated samples are deficient in grammatical quality and semantic consistency, which impairs the effectiveness of adversarial training.
To address these problems, we introduce a novel, effective procedure for instead adversarial training with only clean data. Our procedure, distribution shift risk minimization (DSRM), estimates the adversarial loss by perturbing the input data's probability distribution rather than their embeddings. This formulation results in a robust model that minimizes the expected global loss under adversarial attacks. Our approach requires zero adversarial samples for training and reduces time consumption by up to 70\% compared to current best-performing adversarial training methods.
Experiments demonstrate that DSRM considerably improves BERT's resistance to textual adversarial attacks and achieves state-of-the-art robust accuracy on various benchmarks. 

\end{abstract}

\section{Introduction}
Despite their impressive performance on various NLP tasks, deep neural networks (DNNs), like BERT \cite{devlin2019bert}, are highly vulnerable to adversarial exemplars, which arise by adding imperceptible perturbations among natural samples under semantic and syntactic constraints \cite{zeng2021openattack, lin2021using}. Such vulnerability of DNNs has attracted extensive attention in enhancing defence techniques against adversarial examples \cite{li2021searching, xi2022efficient}, where the adversarial training approach (AT) \cite{goodfellow2014explaining} is empirically one of the best-performing algorithms to train networks robust to adversarial perturbations \cite{uesato2018adversarial,athalye2018obfuscated}. 
Formally, adversarial training attempts to solve the following min-max problem under loss function $L$:
\begin{equation*}
    \min _{\boldsymbol{\theta} \in \Theta} \underbrace{\mathbb{E}_{(\boldsymbol{x}, y) \sim \mathcal{P}_0} { \overbrace{ \color{red}{\max _{\|\boldsymbol{\delta}\|_p \leqslant  \varepsilon} L(\boldsymbol{\theta}, \boldsymbol{x}+\boldsymbol{\delta}, y)} }^{\text{Adversarial Samples (AT)}}}}_{\text{Distribution Shift (Ours)}},
\end{equation*}
where $\boldsymbol{\theta}\in \Theta$ are the model parameters, and $(\boldsymbol{x}, y)$ denotes the input data and label, which follow the joint distribution $\mathcal{P}_0$. The curly brackets show the difference in research focus between our approach and vanilla adversarial training.

Due to the non-convexity of neural networks, finding the analytic solution to the above inner maximization (marked in red) is very difficult \cite{wang2021convergence}. 
The most common approach is to estimate the adversarial loss from the results of several gradient ascents, such as PGD \cite{madry2018towards} and FreeLB \cite{zhu2019freelb}. \citet{li2021token} and \citet{zhu2022improving} generate meaningful sentences by restricting such perturbations to the discrete token embedding space, achieving competitive robustness with better interpretability \cite{shreya2022survey}.

However, the impressive performance in adversarial training comes at the cost of excessive computational consumption, which makes it infeasible for large-scale NLP tasks \cite{andriushchenko2020understanding}. For example, FreeLB++ \cite{li2021searching}, which increases the perturbation intensity of the FreeLB algorithm to serve as one of the state-of-the-art methods, achieves optimal performance with nearly 15 times the training time. Moreover, the adversarial samples generated by the aforementioned methods exhibit poor grammatical quality, which is unreasonable in the real world when being manually reviewed \cite{hauser2021bert, chiang2022far}.  
Some works attempt to speed up the training procedure by obtaining cheaper adversarial samples \cite{wong2019fast} or generating diverse adversarial samples at a negligible additional cost \cite{shafahi2019adversarial}. However, they still require a complex process for adversarial samples and suffer performance degradation in robustness. 

In this work, from another perspective of the overall distribution rather than the individual adversarial samples, we ask the following question: \textit{Can we directly estimate and optimize the expectation of the adversarial loss without computing specific perturbed samples, thus circumventing the above-mentioned problems in adversarial training?}

DSRM formalize the distribution distance between clean and adversarial samples to answer the question. 
Our methodology interprets the generation of adversarial samples as an additional sampling process on the representation space, whose probability density is not uniformly distributed like clean samples. Adversarial samples with higher loss are maximum points in more neighbourhoods and possess a higher probability of being generated. 
We subsequently proved that the intensity of adversarial perturbations naturally bound the Wasserstein distance between these two distributions.
Based on this observation, we propose an upper bound for the adversarial loss, which can be effectively estimated only using the clean training data. By optimizing this upper bound, we can obtain the benefits of adversarial training without computing adversarial samples. In particular, we make the following contributions: 

\begin{itemize}
     \item We propose DSRM, a novel procedure that 
     transforms the training data to a specific distribution to obtain an upper bound on the adversarial loss. Our \textit{codes}\footnote{\url{https://github.com/SleepThroughDifficulties/DSRM}} are publicly available.
     \item We illustrate the validity of our framework with rigorous proofs and provide a practical algorithm based on DSRM, which trains models adversarially without constructing adversarial data.
     \item Through empirical studies on numerous NLP tasks, we show that DSRM significantly improves the adversarial robustness of the language model compared to classical adversarial training methods. In addition, we demonstrate our method's superiority in training speed, which is approximately twice as fast as the vanilla PGD algorithm.
 \end{itemize}

\section{Related Work}
\subsection{Adversarial Training}
% Adversarial training (AT) is a well-accepted method for constructing robust neural networks.
\citet{goodfellow2014explaining} first proposed to generate adversarial samples and utilize them for training. 
Subsequently, the PGD algorithm \cite{madry2018towards} exploits multi-step gradient ascent to search for the optimal perturbations, refining adversarial training into an effective defence technique.
% \cite{qin2019adversarial}. 
Some other works tailored training algorithms for NLP fields to ensure that the adversarial samples have actual sentences. They craft perturbation by replacing words under the guidance of semantic consistency \cite{li2020bert} or token similarity in the embedding space \cite{li2021token}.
However, these algorithms are computationally expensive and trigger explorations to improve training efficiency \cite{zhang2019you}. 
The FreeAT \cite{shafahi2019adversarial} and FreeLB \cite{zhu2019freelb} 
% construct multiple adversarial samples simultaneously in one gradient ascent step to obtain acceleration effects. 
attempt to simplify the computation of gradients to obtain acceleration effects, which construct multiple adversarial samples simultaneously in one gradient ascent step. 
Our DSRM approach is orthogonal to these acceleration techniques as we conduct gradient ascent over the data distribution rather than the input space.

\subsection{Textual Adversarial Samples}
Gradient-based algorithms confront a major challenge in NLP: the texts are discrete, so gradients cannot be directly applied to discrete tokens. \citet{zhu2019freelb} conducts adversarial training by restricting perturbation to the embedding space, which is less interpretable due to the lack of adversarial texts. Some works address this problem by searching for substitution that is similar to gradient-based perturbation \cite{cheng2020advaug, li2021token}. Such substitution strategies can combine with additional rules, such as synonym dictionaries or language models to detect the semantic consistency of adversarial samples \cite{si2021better, zhou2021defense}. 
% These methods help to improve the robustness of NLP models \cite{si2021better, zhou2021defense}. 
However, recent works observe that adversarial samples generated by these substitution methods are often filled with syntactic errors and do not preserve the semantics of the original inputs \cite{hauser2021bert, chiang2022far}. \citet{wang2022distinguishing}
constructs discriminative models to select beneficial adversarial samples, such a procedure further increases the time consumption of adversarial training. In this paper, we propose to estimate the global adversarial loss with only clean data, thus circumventing the defects in adversarial sample generation and selection.
\section{Methodology}
In this section, we first introduce our distribution shift risk minimization (DSRM) objective, a novel upper bound estimation for robust optimization, and subsequently, how to optimize the model parameters under DSRM.

Throughout our paper, we denote vectors as $\boldsymbol{a}$, sets as $\mathcal{A}$, probability distributions as $\mathcal{P}$, and definition as $\triangleq$. Specificly, we denote an all-1 vector of length $b$ as $\vec{1}_{1 \times b}$.
Considering a model parameterized by $\boldsymbol{\theta} \in \Theta$, the per-data loss function is denoted as $L(\boldsymbol{\theta}, \boldsymbol{x}, y): \Theta \times \mathcal{X} \times \mathcal{Y} \rightarrow \mathbb{R}_{+}$. Observing only the training set $\mathcal{S}_t$, the goal of model training is to select model parameters $\boldsymbol{\theta}$ that are robust to adversarial attacks.

\subsection{Adversarial Loss Estimation by Distribution Shift}
We initiate our derivation with vanilla PGD objective \cite{madry2017towards}. Formally, PGD attempts to solve the following min-max problem:
\begin{align*}
    \min_{\boldsymbol{\theta} \in \Theta} \rho(\boldsymbol{\theta}) \triangleq  \mathbb{E}_{(\boldsymbol{x}, y) \sim \mathcal{P}_0} { {\max _{\|\boldsymbol{\delta}\|_p \leqslant  \varepsilon} L(\boldsymbol{\theta}, \boldsymbol{x}+\boldsymbol{\delta}, y)}},
\end{align*}
where $\boldsymbol{\theta}\in \Theta$ are the model parameters, and $(\boldsymbol{x}, y)$ denotes the input data and label, which follow the joint distribution $\mathcal{P}_0$. 

Instead of computing the optimal perturbation for each data point, we directly study the $\rho(\boldsymbol{\theta})$ from the data distribution perspective. During the training process of PGD, each input $\boldsymbol{x}$ corresponds to an implicit adversarial sample. We describe such mapping relationship with a transformation functions $f:X\ \times\   Y\rightarrow{X}$ as:
\begin{align}\label{fomula:1}
f_{\varepsilon, \boldsymbol{\theta}}(\boldsymbol{x}, y) \triangleq \boldsymbol{x}+\arg\max_{\left\{\boldsymbol{\delta}:\|\boldsymbol{\delta}\|_{p} \leq \varepsilon\right\}} L(\boldsymbol{\theta}, \boldsymbol{x}+\boldsymbol{\delta}, y).
\end{align}

The existence of $f_{\varepsilon, \boldsymbol{\theta}}(\boldsymbol{x}, y)$ can be guaranteed due to the continuity of the loss function $L(\boldsymbol{\theta}, \boldsymbol{x}+\boldsymbol{\delta}, y)$. Then the training objective $\rho(\boldsymbol{\theta})$ can be denoted as:
\begin{align}
    \rho(\boldsymbol{\theta}) &=  \mathbb{E}_{(\boldsymbol{x}, y) \sim \mathcal{P}_0} \ { L(\boldsymbol{\theta}, f_{\varepsilon, \boldsymbol{\theta}}(\boldsymbol{x}, y), y)} \\
    &= \mathbb{E}_{(\boldsymbol{x}, y) \sim \mathcal{P}_f} \ { L(\boldsymbol{\theta}, \boldsymbol{x}, y)},
    \label{fomula:2}
\end{align}
where $\mathcal{P}_f$ denotes the distribution of $f_{\varepsilon, \boldsymbol{\theta}}(\boldsymbol{x}, y)$. Eq. \ref{fomula:2} omits the perturbation $\boldsymbol{\delta}$ by introducing $\mathcal{P}_f$, and directly approximates the robust optimization loss. However, the accurate distribution is intractable due to the non-convex nature of neural networks. We, therefore, constrain the above distribution shift (i.e., from $\mathcal{P}_0$ to $\mathcal{P}_f$) with Wasserstein distance.
\newtheorem{theorem}{Theorem}[section]
\newtheorem{definition}{Definition}[section]
\newtheorem{lemma}{Lemma}[section]

\begin{lemma}[]\label{lemma:1}
Let $\mathrm{W}_{p}\left(\mathcal{P}, \mathcal{Q}\right)$ denotes the $p$-th Wasserstein distance between $\mathcal{P}$ and $ \mathcal{Q}$ \cite{peyre2019computational}. $\mathcal{P}_0$ and $\mathcal{P}_f$ are the respective distributions of clean and perturbed samples. The p-norm of perturbation $\delta$ is constrained by $\|\boldsymbol{\delta}\|_{p} \leq \varepsilon$, then the distribution shift in Eq. \ref{fomula:2} is bounded by:
\begin{equation*}
\mathrm{W}_{p}\left(\mathcal{P}_0, \mathcal{P}_f\right) \leq \varepsilon
\end{equation*}
\end{lemma}

\begin{proof}\let\qed\relax
With Eq. \ref{fomula:1}, we have:
% $\left\|\boldsymbol{x}-f_{\varepsilon, \boldsymbol{\theta}}(\boldsymbol{x})\right\|_{p}\leq \varepsilon$
\begin{equation*}\resizebox{0.96\hsize}{!}{$\begin{aligned}
\mathrm{W}_{p}\left(\mathcal{P}_0, \mathcal{P}_f\right)&\triangleq\left(\inf _{\pi \in \Pi(\mathcal{P}_0, \mathcal{P}_f)} \mathbb{E}_{(u, \boldsymbol{v}) \sim \pi}\left[\|\boldsymbol{u}-\boldsymbol{v}\|_p^p\right]\right)^{\frac{1}{p}}\\
&\leq \left(\mathbb{E}_{(\boldsymbol{x}, y) \sim \mathcal{P}_0}\left[\left\|\boldsymbol{x}-f_{\varepsilon, \boldsymbol{\theta}}(\boldsymbol{x}, y)\right\|_{p}^p\right]\right)^{\frac{1}{p}}\\
&\leq \varepsilon.
\end{aligned}$}
\end{equation*}
\end{proof}
Lemma. \ref{lemma:1} ensures that for bounded perturbation strengths, the distribution shift between the original and virtual adversarial samples is limited, and we consequently define our Distribution Shift Risk Minimization (DSRM) objective as follows:
\begin{definition}[DSRM]\label{def: dsrm}
Giving $(\boldsymbol{x}, y)\sim \mathcal{P}_0$, loss function $L$ and model parameters $\theta$, the DSRM aiming to minimize the worst-case loss $\rho_{DS} (\boldsymbol{\theta})$ under distributional perturbations with intensity limited to $\varepsilon$, that is:
% $\min_{\boldsymbol{\theta} \in \Theta} \rho_{DS}(\theta)$, where
\begin{align}\label{fomula:4}
    \min_{\boldsymbol{\theta} \in \Theta}
    \rho_{DS} (\boldsymbol{\theta}) \triangleq   {\max _{\mathrm{W}_{p}\left(\mathcal{P}_0,\  \mathcal{P}_t\right) \leqslant  \varepsilon} \mathbb{E}_{(\boldsymbol{x}, y) \sim \mathcal{P}_t} {  L(\boldsymbol{\theta}, \boldsymbol{x}, y)}}.
\end{align}
\end{definition}

Noticing that there always satisfies:
\begin{align*}
    \rho (\boldsymbol{\theta}) \leq \rho_{DS} (\boldsymbol{\theta}),
\end{align*}
we subsequently optimize the upper bound $\rho_{DS} (\theta)$ for adversarial training.

\subsection{Distribution Shift Adversarial Training}
In definition \ref{def: dsrm}, we propose DSRM, a new adversarial training objective from the perspective of distribution shift. 
We now discuss how to optimize the model parameters with a finite training set $\mathcal{S} \triangleq \cup_{i=1}^n\left\{\left(\boldsymbol{x}_i, \boldsymbol{y}_i\right)\right\}$. We first introduce the empirical estimation of Eq. \ref{fomula:4} as follows:
% \begin{small}
\begin{align*}
    \rho_{DS} (\theta) \approx  {\max _{\mathrm{W}_{p}\left(\mathcal{P}_0,\  \mathcal{P}_t\right) \leqslant  \varepsilon} \sum_{i=1}^n {\mathcal{P}_t (\boldsymbol{x}_i)  L(\boldsymbol{\theta}, \boldsymbol{x}_i, y_i)}},
\end{align*}
% \end{small}
% Let $\Delta\mathcal{P}$ to denote the change between worst-case distribution, 
% where the unperturbed distribution $\mathcal{P}_0$ is a uniform distribution. For the purpose of simplicity,
where $\mathcal{P}_0$ is the unperturbed distribution. In vanilla training procedure, all training data are weighted as $\frac{1}{n}$, where $n$ is the value of training batch size. We therefore model $\mathcal{P}_0$ as a uniform distribution. For the purpose of simplicity, we use $L_S(\boldsymbol{\theta}, \mathcal{P}_t)$ to denote the inner maximization term, that is:
\begin{align}
    \rho_{DS} (\boldsymbol{\theta}) \approx  {\max _{\mathrm{W}_{p}\left(\mathcal{P}_0,\  \mathcal{P}_t\right) \leqslant  \varepsilon} L_S(\boldsymbol{\theta}, \mathcal{P}_t)}.
\end{align}

Suppose the worst-case distribution is $\mathcal{P}_f$.
To make explicit our distribution shift term, we rewrite the right-hand side of the equation above as:
\begin{align*}
    L_S(\boldsymbol{\theta},\mathcal{P}_0) + \left[ {\sum_{i=1}^n {\left(\mathcal{P}_f (\boldsymbol{x}_i) - \frac{1}{n} \right)  L(\boldsymbol{\theta}, \boldsymbol{x}_i, y_i)}}\right],
\end{align*}
where $L_S(\boldsymbol{\theta},\mathcal{P}_0) \triangleq \frac{1}{n} \sum_{i=1}^n L(\boldsymbol{\theta}, \boldsymbol{x}_i, y_i)$ are the empirical risk of training sets. 
The term in square brackets captures the sensitivity of $\rho_{DS} (\theta)$ at $\mathcal{P}_f$, measuring how quickly the empirical loss increase when transforming training samples to different weights. This term can be denoted as $ L_S(\boldsymbol{\theta}, \mathcal{P}_f - \mathcal{P}_0)$. 
Since the training set is finite, the probability distribution over all samples can be simplified to a vector, let $\boldsymbol{P}_f = \left[ \mathcal{P}_f (\boldsymbol{x}_1), \mathcal{P}_f (\boldsymbol{x}_2),...,\mathcal{P}_f (\boldsymbol{x}_{n})  \right]$, and $\boldsymbol{L} = \left[  L(\boldsymbol{\theta}, \boldsymbol{x}_1, y_1), L(\boldsymbol{\theta}, \boldsymbol{x}_2, y_2),...,L(\boldsymbol{\theta}, \boldsymbol{x}_n, y_n)  \right]$, we have:
\begin{align}
    L_S(\boldsymbol{\theta}, \mathcal{P}_f - \mathcal{P}_0) = \left(\boldsymbol{P}_f - \frac{1}{n}\right)\boldsymbol{L}^{T}.
\end{align}

% and now discuss how to estimate it over successive training procedures.
In order to minimize the $ L_S(\boldsymbol{\theta}, \mathcal{P}_f)$, we first derive an approximation to the inner maximization of DSRM. We approximate the inner maximization problem via a first-order Taylor expansion of $\rho_{DS} (\theta)$ w.r.t $\mathcal{P}_f$ around $\mathcal{P}_0$, we obtain the estimation as follows:
\begin{equation}\resizebox{0.96\hsize}{!}{$
\begin{aligned} \label{fomula:8}
     \mathcal{P}_f &= \arg {\max _{\mathrm{W}_{p}\left(\mathcal{P}_0,\  \mathcal{P}_t\right) \leqslant  \varepsilon} L_S(\boldsymbol{\theta}, \mathcal{P}_t)}\\
     &= \arg {\max _{\mathrm{W}_{p}\left(\mathcal{P}_0,\  \mathcal{P}_t\right) \leqslant  \varepsilon}  \left[L_S(\boldsymbol{\theta}, \mathcal{P}_t)-L_S(\boldsymbol{\theta}, \mathcal{P}_0)\right]}\\
     &\approx \arg {\max _{\mathrm{W}_{p}\left(\mathcal{P}_0,\  \mathcal{P}_t\right) \leqslant  \varepsilon} \left[ \left( \mathcal{P}_t - \mathcal{P}_0    \right)^T\nabla_{\mathcal{P}_t}  L_S(\boldsymbol{\theta}, \mathcal{P}_0)\right]}.
\end{aligned}$}
\end{equation}

By Eq. \ref{fomula:8}, the value $\mathcal{P}_f$ that exactly solves this approximation can be given by its dual problem. For experimental convenience, here we only focus on and present one of the special cases, that the metric used in $\mathrm{W}_{p}\left(\mathcal{P}_0,\  \mathcal{P}_t\right)$ treats all data pairs equally. We empirically demonstrate that such approximations can achieve promising performance in the next section.
In turn, the solution of $\mathcal{P}_f$ can be denoted as:
\begin{equation}\resizebox{0.95\hsize}{!}{$
\begin{aligned}
    \mathcal{P}_f^{*} = \varepsilon \ \nabla_{\mathcal{P}_t}  L_S(\boldsymbol{\theta}, \mathcal{P}_0)\ /\ 
    \|\nabla_{\mathcal{P}_t} L_S(\boldsymbol{\theta}, \mathcal{P}_0)\| + \mathcal{P}_0.
\end{aligned}$}
\end{equation}
Substituting the equation into Eq. \ref{fomula:4} and differentiating the DSRM objective, we then have:
\begin{equation}\resizebox{0.97\hsize}{!}{$
\begin{aligned}
    &\nabla_{\boldsymbol{\theta}} ( \rho_{DS}(\boldsymbol{\theta})) \approx 
    \nabla_{\boldsymbol{\theta}}
    L_S\left(\boldsymbol{\theta}, \mathcal{P}_f^{*}\right) \\
    =&\nabla_{\boldsymbol{\theta}}\left[L_S\left(\boldsymbol{\theta}, \mathcal{P}_0\right) +  \left(\mathcal{P}_f^{*}-\mathcal{P}_0 \right) \nabla_{\mathcal{P}_t}L_S\left(\boldsymbol{\theta}, \mathcal{P}_t\right) |_{\mathcal{P}_f^{*}}\right].
\end{aligned}$}
\end{equation}

Though this approximation to $\nabla_{\boldsymbol{\theta}} ( \rho_{DS}(\theta))$ requires a potential second-order differentiation (the influence of weight perturbations on the loss of DSRM), 
they can be decomposed into a multi-step process, which is tractable with an automatic meta-learning framework. In our experiments, 
we use the Higher \footnote{\url{https://github.com/facebookresearch/higher.git}.} package for differential to the sample weight. 

To summarize, we first update the parameters for one step under the original data distribution $\mathcal{P}_0$, and compute the empirical loss on a previously divided validation set, which requires an additional set of forward processes with the updated parameters. Later, we differentiate validation loss to the weights of the input samples to obtain the worst-case perturbation and re-update the parameters with our distribution shift loss function.
Our detailed algorithm implementation is shown in Algorithm \ref{alg:Framwork}.

\begin{algorithm}[ht]
\caption{ Framework of Our DSRM.}
\label{alg:Framwork}
\KwIn{Training set $\mathcal{S}_t$, Validate set $\mathcal{S}_v$, Loss function $L$, Batch size $b$, Pre-trained model parameters $\theta$, Batch sample weights $\boldsymbol{w}$, Optimizer $\delta$, Perturbation size $\eta$.}
\KwOut{Model trained with DSRM} 
Initialize: Virtual Model $\theta^{\star}=\theta$, Step $n=0$\;
\While{\textit{not converged}}{
    $\boldsymbol{w} = \vec{1}_{1 \times b}$ \;
    Sample batch train data from $\mathcal{S}_t$: $\mathcal{B}_t=\{\left(\boldsymbol{x}_i, \boldsymbol{y}_i\right)\}_{i=1}^b$\;
    Compute gradient of the batch's empirical loss: $\nabla_{\theta_{n}^{\star}} \boldsymbol{w}^T L\left( \theta_{n}^{\star},  \mathcal{B}_t \right)$\;
    Update virtual model: $\theta_{n+1}^{\star} = \theta_{n}^{\star} - \delta\left(\nabla_{\theta_{n}^{\star}} \boldsymbol{w}^T L\left( \theta_{n}^{\star},  \mathcal{B}_t \right)\right)$\;
    Sample valid data from $\mathcal{S}_v$: $\mathcal{B}_v=\{\left(\boldsymbol{x}_j, \boldsymbol{y}_j\right)\}_{j=1}^{2b}$\;
    Compute gradient of global loss to sample weights: $g = \nabla_{\boldsymbol{w}} L\left( \theta_{n+1}^{\star}, \mathcal{B}_v \right)$\;
    Update $\boldsymbol{w}$ with gradient $g$: $\boldsymbol{w}_n = \boldsymbol{w} + \eta g$\;
    Compute gradient of the DSRM objective: $\nabla_{\theta_n} \boldsymbol{w}_n^T L\left(  \theta_{n}, \mathcal{B}_t \right)$\;
    Update model parameters: $\theta_{n+1} = \theta_{n} - \delta\left(\nabla_{\theta_n} \boldsymbol{w}_n^T L\left(  \theta_{n}, \mathcal{B}_t \right)\right)$\;
    $n = n + 1$
}
\Return $\theta_{n}$
\end{algorithm}

\section{Experiments}
In this section, we comprehensively analyse DSRM versus other adversarial training methods in three evaluation settings for three tasks.

\subsection{Datasets and Backbone Model}

We evaluate our proposed method mainly on the four most commonly used classification tasks for adversarial defence, including SST-2 \cite{socher2013recursive}, IMDB \cite{maas2011learning}, AG NEWS \cite{zhang2015character} and QNLI \cite{wang2018glue}. The statistics of these involved benchmark datasets are summarised in Appendix \ref{app: dataset}. We take the BERT-base model (12 transformer layers, 12 attention heads, and 110M parameters in total) as the backbone model, and follow the BERT implementations in \cite{devlin2019bert}.
% \usepackage{booktabs}

\subsection{Evaluation Settings}
We refer to the setup of previous state-of-the-art works \cite{liu2022flooding, xi2022efficient} to verify the robustness of the model. The pre-trained model is finetuned with different defence methods on various datasets and saves the best three checkpoints. We then test the defensive capabilities of the saved checkpoint via TextAttack \cite{morris2020textattack} and report the mean value as the result of the robustness evaluation experiments.

Three well-received textual attack methods are leveraged in our experiments. TextBugger \cite{li2018textbugger} identify the critical words of the target model and repeatedly replace them with synonyms until the model's predictions are changed. TextFooler \cite{jin2020bert} similarly filter the keywords in the sentences and select an optimal perturbation from various generated candidates. BERTAttack \cite{li2020bert} applies BERT to maintain semantic consistency and generate substitutions for vulnerable words detected in the input. 

For all attack methods, we introduce four metrics to measure BERT's resistance to adversarial attacks under different defence algorithms.
\textbf{Clean accuracy (Clean\%)} refers to the model's test accuracy on the clean dataset. \textbf{Accurucy under attack (Aua\%)} refers to the model's prediction accuracy with the adversarial data generated by specific attack methods. \textbf{Attack success rate (Suc\%)} measures the ratio of the number of texts successfully scrambled by a specific attack method to the number of all texts involved. \textbf{Number of Queries (\#Query)} refers to the average attempts the attacker queries the target model. The larger the number is, the more complex the model is to be attacked.

\subsection{Baseline Methods}
Since our method is based on the adversarial training objective, we mainly compare it with previous adversarial training algorithms. In addition, to refine the demonstration of the effectiveness of our method, we also introduce two non-adversarial training methods (InfoBERT and Flooding-X) from current state-of-the-art works.

\begin{table*}[ht]
\resizebox{\textwidth}{!}{%
\begin{tabular}{c|l|c|ccc|ccc|ccc}
\hline\hline
 & \multicolumn{1}{c|}{} &  & \multicolumn{3}{c|}{\textbf{TextFooler}} & \multicolumn{3}{c|}{\textbf{BERT-Attack}} & \multicolumn{3}{c}{\textbf{TextBugger}} \\ \cline{4-12} 
\multirow{-2}{*}{\textbf{Datasets}} & \multicolumn{1}{c|}{\multirow{-2}{*}{\textbf{Methods}}} & \multirow{-2}{*}{\textbf{Clean\%}} & Aua\% & Suc\% & \#Query & Aua\% & Suc\% & \#Query & Aua\% & Suc\% & \#Query \\ \hline
 & Fine-tune & 93.1 & 5.7 & 94.0 & 89.3 & 5.9 & 93.4 & 108.9 & 28.2 & 68.7 & 49.2 \\
 & PGD$^\dag$ & 92.8 & 8.3 & 90.7 & 94.6 & 8.7 & 90.5 & 117.7 & 31.5 & 65.2 & 53.3 \\
 & FreeLB$^\dag$ & 93.6 & 8.5 & 91.4 & 95.4 & 9.3 & 90.2 & 118.7 & 31.8 & 64.7 & 50.2 \\
 & FreeLB++$^\dag$ & 92.9 & 14.3 & 84.8 & 118.2 & 11.7 & 87.4 & 139.9 & 37.4 & 61.2 & 52.3 \\
 & TAVAT$^\dag$ & 93.0 & 12.5 & 85.3 & 121.7 & 11.6 & 85.3 & 129.0 & 29.3 & 67.2 & 48.6 \\  
 & InfoBERT$^\ddag$ & 92.9 & 12.5 & 85.1 & 122.8 & 13.4 & 83.6 & 133.3 & 33.4 & 63.8 & 50.9 \\
 & Flooding-X$^\ddag$ & \textbf{93.1} & 28.4 & 67.5 & 149.6 & 25.3 & 70.7 & 192.4 & 41.9 & 58.3 & 62.5 \\ \cline{2-12} 
\multirow{-8}{*}{\textbf{SST-2}} & \cellcolor[HTML]{E0E0E0}DSRM(ours) & \cellcolor[HTML]{E0E0E0}91.5 & \cellcolor[HTML]{E0E0E0}\textbf{32.8} & \cellcolor[HTML]{E0E0E0}\textbf{65.1} & \cellcolor[HTML]{E0E0E0}\textbf{153.6} & \cellcolor[HTML]{E0E0E0}\textbf{27.2} & \cellcolor[HTML]{E0E0E0}\textbf{69.1} & \cellcolor[HTML]{E0E0E0}\textbf{201.5} & \cellcolor[HTML]{E0E0E0}\textbf{44.2} & \cellcolor[HTML]{E0E0E0}\textbf{51.4} & \cellcolor[HTML]{E0E0E0}\textbf{88.6} \\ \hline
 & Fine-tune & 90.6 & 5.8 & 94.2 & 161.9 & 3.5 & 96.1 & 216.5 & 10.9 & 88.0 & 98.4 \\
 & PGD$^\dag$ & 90.6 & 14.3 & 81.2 & 201.6 & 17.3 & 80.6 & 268.9 & 27.9 & 67.8 & 134.6 \\
 & FreeLB$^\dag$ & 90.7 & 12.8 & 85.3 & 189.4 & \textbf{21.4} & 76.8 & \textbf{324.2} & 29.8 & 69.3 & 143.9 \\
 & FreeLB++$^\dag$ & \textbf{91.1} & 16.4 & 81.4 & 193.7 & 20.7 & 77.0 & 301.7 & 30.2 & 66.7 & 150.1 \\  
 & InfoBERT$^\ddag$ & 90.4 & 18.0 & 82.5 & 212.9 & 13.1 & 85.8 & 270.2 & 15.4 & 83.9 & 127.9 \\
 & Flooding-X$^\ddag$ & 90.8 & 25.6 & 71.3 & 232.7 & 18.7 & 79.2 & 294.6 & 29.4 & 67.5 & 137.1 \\ \cline{2-12} 
\multirow{-7}{*}{\textbf{QNLI}} & \cellcolor[HTML]{E0E0E0}DSRM(ours) & \cellcolor[HTML]{E0E0E0}90.1 & \cellcolor[HTML]{E0E0E0}\textbf{27.6} & \cellcolor[HTML]{E0E0E0}\textbf{65.4} & \cellcolor[HTML]{E0E0E0}\textbf{247.2} & \cellcolor[HTML]{E0E0E0}20.4 & \cellcolor[HTML]{E0E0E0}\textbf{76.7} & \cellcolor[HTML]{E0E0E0}312.4 & \cellcolor[HTML]{E0E0E0}\textbf{37.1} & \cellcolor[HTML]{E0E0E0}\textbf{59.2} & \cellcolor[HTML]{E0E0E0}\textbf{176.3} \\ \hline
 & Fine-tune & 92.1 & 10.3 & 88.8 & 922.4 & 5.3 & 94.3 & 1187.0 & 15.8 & 83.7 & 695.2 \\
 & PGD$^\dag$ & 93.2 & 26.0 & 72.1 & 1562.8 & 21.0 & 77.6 & 2114.6 & 41.6 & 53.2 & 905.8 \\
 & FreeLB$^\dag$ & 93.2 & 35.0 & 62.7 & 1736.9 & 29.0 & 68.4 & 2588.8 & 53.0 & 44.2 & 1110.9 \\
 & FreeLB++$^\dag$ & 93.2 & 45.3 & 51.0 & 1895.3 & 39.9 & 56.9 & 2732.5 & 42.9 & 54.6 & 1094.0 \\
 & TAVAT$^\dag$ & 92.7 & 27.6 & 71.9 & 1405.8 & 23.1 & 75.1 & 2244.8 & 54.1 & 44.1 & 1022.6 \\ 
 & InfoBERT$^\ddag$ & 93.3 & 49.6 & 49.1 & 1932.3 & 47.2 & 51.3 & 3088.8 & 53.8 & 44.7 & 1070.4 \\
 & Flooding-X$^\ddag$ & 93.4 & 45.5 & 53.5 & 2015.4 & 37.3 & 60.8 & 2448.7 & 62.3 & 35.8 & 1187.9 \\ \cline{2-12} 
\multirow{-8}{*}{\textbf{IMDB}} & \cellcolor[HTML]{E0E0E0}DSRM(ours) & \cellcolor[HTML]{E0E0E0}\textbf{93.4} & \cellcolor[HTML]{E0E0E0}\textbf{56.3} & \cellcolor[HTML]{E0E0E0}\textbf{39.0} & \cellcolor[HTML]{E0E0E0}\textbf{2215.3} & \cellcolor[HTML]{E0E0E0}\textbf{54.1} & \cellcolor[HTML]{E0E0E0}\textbf{41.2} & \cellcolor[HTML]{E0E0E0}\textbf{3309.8} & \cellcolor[HTML]{E0E0E0}\textbf{67.2} & \cellcolor[HTML]{E0E0E0}\textbf{28.9} & \cellcolor[HTML]{E0E0E0}\textbf{1207.7} \\ \hline
 & Fine-tune & 93.9 & 28.6 & 69.9 & 383.3 & 17.6 & 81.2 & 556.0 & 45.2 & 53.4 & 192.5 \\
 & PGD$^\dag$ & 94.5 & 36.8 & 68.2 & 414.9 & 21.6 & 77.1 & 616.1 & 56.4 & 41.9 & 201.8 \\
 & FreeLB$^\dag$ & 94.7 & 34.8 & 63.4 & 408.5 & 20.4 & 73.8 & 596.2 & 54.2 & 43.0 & 210.3 \\
 & FreeLB++$^\dag$ & 94.9 & 51.5 & 46.0 & 439.1 & 41.8 & 56.2 & 676.4 & 55.9 & 41.4 & 265.4 \\
 & TAVAT$^\dag$ & \textbf{95.2} & 31.8 & 66.5 & 369.9 & 35.0 & 62.5 & 634.9 & 54.2 & 43.9 & 231.2 \\  
 & InfoBERT$^\ddag$ & 94.5 & 33.8 & 65.1 & 395.6 & 23.4 & 75.3 & 618.9 & 49.6 & 47.7 & 194.1 \\
 & Flooding-X$^\ddag$ & 94.8 & 42.4 & 54.9 & 421.4 & 27.4 & 71.0 & 590.3 & 62.2 & 34.0 & 272.5 \\ \cline{2-12} 
\multirow{-8}{*}{\textbf{AG NEWS}} & \cellcolor[HTML]{E0E0E0}DSRM(ours) & \cellcolor[HTML]{E0E0E0}93.5 & \cellcolor[HTML]{E0E0E0}\textbf{62.9} & \cellcolor[HTML]{E0E0E0}\textbf{31.4} & \cellcolor[HTML]{E0E0E0}\textbf{495.0} & \cellcolor[HTML]{E0E0E0}\textbf{58.6} & \cellcolor[HTML]{E0E0E0}\textbf{36.1} & \cellcolor[HTML]{E0E0E0}\textbf{797.7} & \cellcolor[HTML]{E0E0E0}\textbf{69.4} & \cellcolor[HTML]{E0E0E0}\textbf{24.8} & \cellcolor[HTML]{E0E0E0}\textbf{294.6} \\ \hline\hline
\end{tabular}%
}
\caption{The experiment results of different defenders on four datasets. The best performance of each metric is marked in \textbf{bold}. Methods labelled by $\dag$ are adversarial training baselines, and methods labelled by $\ddag$ are other SOTA defence baselines. Our method surpasses existing methods by a large margin under almost all tasks.}
\label{tab:2}
\end{table*}

\paragraph*{PGD} Projected gradient descent \cite{madry2018towards} formulates adversarial training algorithms to minimize the empirical loss on adversarial examples.
\paragraph*{FreeLB} FreeLB \cite{zhu2019freelb} generates virtual adversarial samples in the region surrounding the input samples by adding adversarial perturbations to the word embeddings.
\paragraph*{FreeLB++} Based on FreeLB, \citet{li2021searching} discovered that the effectiveness of adversarial training could be improved by scaling up the steps of FreeLB, and proposed FreeLB++, which exhibits the current optimal results in textual adversarial training.
\paragraph*{TAVAT} Token-Aware Virtual Adversarial Training \cite{li2021token} proposed a token-level perturbation vocabulary to constrain adversarial training within a token-level normalization ball.
\paragraph*{InfoBERT} InfoBERT \cite{wang2020infobert} leverages two regularizers based on mutual information, enabling models to explore stable features better.
\paragraph*{Flooding-X} Flooding-X \cite{liu2022flooding} smooth the parameter landscape with Flooding \cite{ishida2020we} to boost model resistance to adversarial perturbations.

\subsection{Implementation Details}
We reproduced the baseline works based on their open-source codes, and the results are competitive relative to what they reported in the paper. The \textbf{Clean\%} is evaluated on the whole test set. 
\textbf{Aua\%}, \textbf{Suc\%} and \textbf{\#Query} are evaluated on the whole test dataset for SST-2, and on 1000 randomly selected samples for the other three datasets. We train our models on NVIDIA RTX 3090 GPUs. Most parameters, such as learning rate and warm-up steps, are consistent with the FreeLB \cite{zhu2019freelb}.
 We train 8 epochs with 3 random seeds for each model on each dataset and report the resulting mean error (or accuracy) on test sets. To reduce the time consumption for calculating the distribution shift risk, for each step we sample 64 sentences (32 for IMDB) from the validation set to estimate our adversarial loss.
More implementation details and hyperparameters can be found in Appendix \ref{app:hyper}.
% DSRM has a single hyperparameter $\varepsilon$ (the constraints on perturbation intensity) which we tune via a grid search over \{0.1,0.2,0.4,0.6,0.8,1.0\}.
\subsection{Experimental Results}

Our analysis of the DSRM approach with other comparative methods against various adversarial attacks is summarized in Table \ref{tab:2}. Our method demonstrates significant improvements in the BERT’s resistance to these attacks, outperforming the baseline defence algorithm on most datasets.

In the SST-2, IMDB, and AG NEWS datasets, DSRM achieved optimal robustness against all three attack algorithms. It is worth noting that the effectiveness of DSRM was more pronounced on the more complex IMDB and AG NEWS datasets, as the estimation of adversarial loss for these tasks is more challenging than for the simpler SST-2 dataset. This phenomenon verifies that our method better estimates the inner maximation problem. In the QNLI dataset, DSRM only fails to win in the BertAttack, but still maintains the lowest attack success rate among all methods, with an Aua\% that is only 1\% lower than that of FreeLB. This difference in performance can be attributed to the varying clean accuracy of the two methods, in which case DSRM misclassifies a small number of samples that are more robust to the attack.

In terms of clean accuracy, our method suffers from a minor degradation on SST-2, QNLI and AGNEWS, which is acceptable as a trade-off in robustness and generalization for adversarial training, and we will further discuss this phenomenon in the next section. On IMDB, our approach achieves the best clean accuracy together with flooding-X. We attribute this gain to the greater complexity of the IMDB dataset, that the aforementioned trade-off appears later to enable DSRM to achieve better performance.

Overall, DSRM performs better than the baseline adversarial training methods by 5 to 20 points on average without using any adversarial examples as training sources. Besides, our approach is more effective for complex datasets and remains the best-performing algorithm on Textfooler and Textbugger, which demonstrates the versatility and effectiveness of DSRM. Our experiments demonstrate that adversarial training methods have a richer potential for constructing robust language models.
\section{Analysis and Discussion}
In this section, we construct supplementary experiments to analyze our DSRM framework further.

\subsection{DSRM Induces Smooth Loss Distribution}
\begin{figure}[htbp]
\centering
 \subfigure[]{
\begin{minipage}[b]{0.225\textwidth}
\includegraphics[width=1\textwidth]{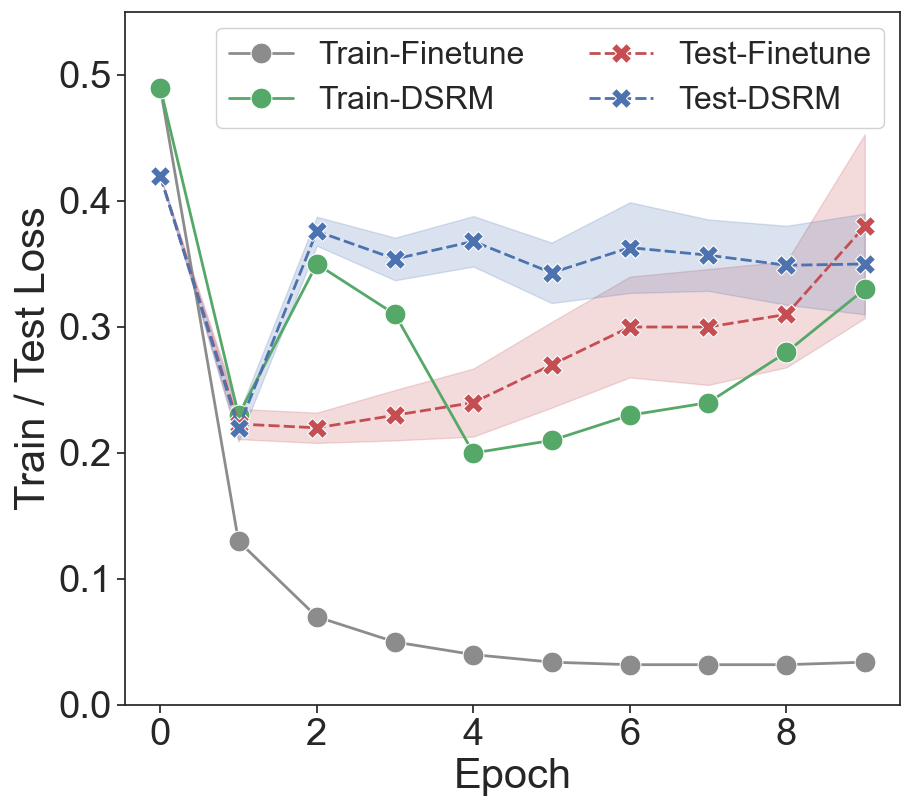}
  \end{minipage}
 }
     \subfigure[]{
      \begin{minipage}[b]{0.225\textwidth}
       \includegraphics[width=1\textwidth]{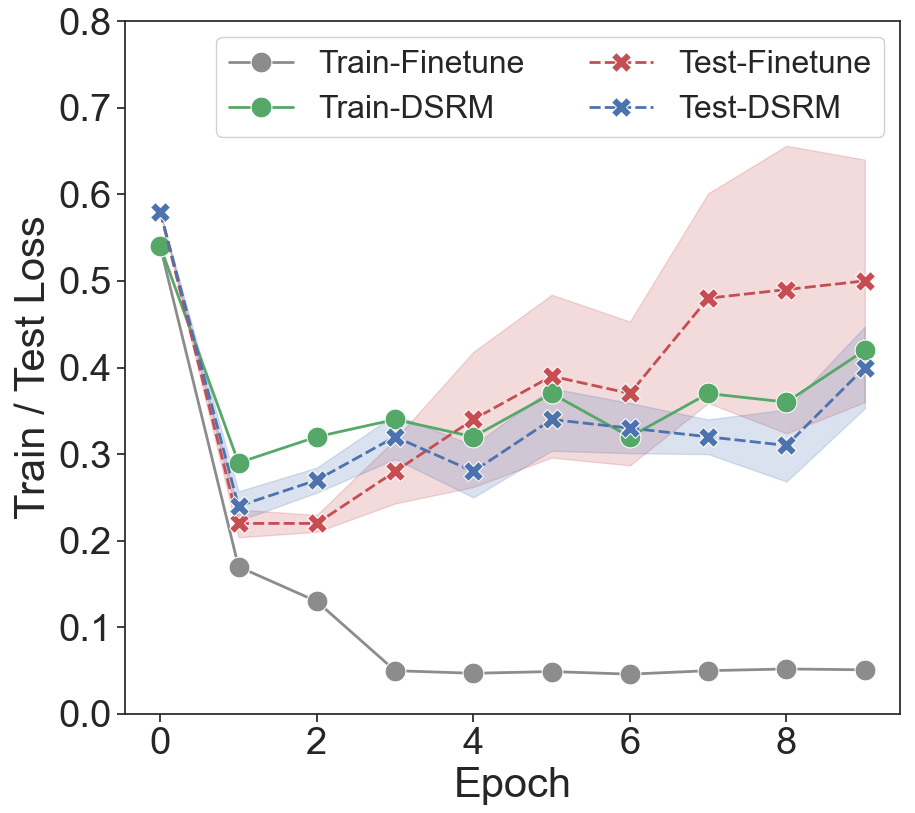}
      \end{minipage}
     }
 \caption{The train/test loss of DSRM and fine-tuning on the SST-2 (a) and IMDB (b) datasets. We report the mean loss on the train and test sets, and variance (marked with shadow) only on the test set. Our method maintains uniform loss distribution and better consistency between training and test data while the fine-tuning overfits quickly after one epoch.}
\vspace{-0.7em}
\label{fig: loss}
\end{figure}

Previous works demonstrate that deep neural networks suffer from overfitting training configurations and memorizing training samples, leading to poor generalization error and vulnerability towards adversarial perturbations \cite{werpachowski2019detecting, rodriguez2021evaluation}. We verify that DSRM mitigates such overfitting problems by implicitly regularizing the loss's smoothness in the input space. Figure \ref{fig: loss} shows the
training/test loss of each BERT epoch trained by DSRM and fine-tuning. Models trained by fine-tuning overfit quickly and suffer persistent performance degradation as the epoch grows. In contrast, the loss curves of our method maintain lower generalization errors with a minor variance of the predicted losses on the test set. This improvement comes from the fact that under the training objective of DSRM, where the model allocates more attention to samples with a higher loss.

\subsection{Effect of Perturbation Intensity}
\begin{figure}[htbp]
\centering
\includegraphics[width=0.42\textwidth]{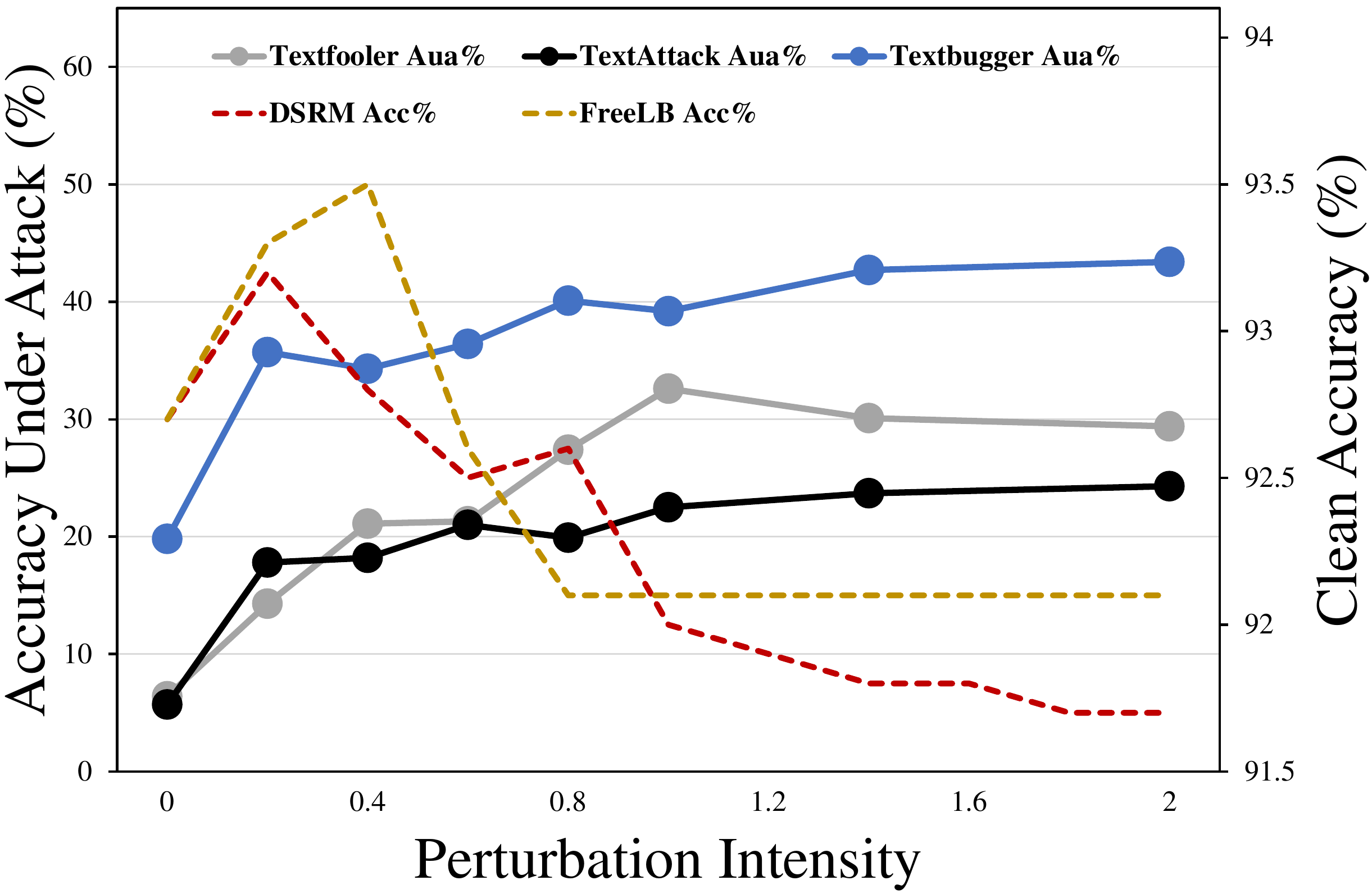}
\caption{Accuracy and Aua\% of BERT trained by DSRM under different perturbations.}
\label{fig: perturb}
\vspace{-0.7em}
\end{figure}

DSRM has a single hyperparameter $\varepsilon$ to control the constraints on perturbation intensity. The extension in the perturbation range brings a better optimization on the defence objective, while the mismatch between the train and test set data distribution may impair the model performance. To further analyze the impact of DSRM on model accuracy and robustness, we conduct a sensitivity analysis of perturbation intensity $\varepsilon$. Figure \ref{fig: perturb} illustrates the variation curve of
performance change for our method on three attack algorithms. 

DSRM improves accuracy and Aua\% when perturbations are moderated ($\leq0.2$), similar to other adversarial training methods. When the perturbation becomes stronger, the model's resistance to adversarial attacks improves notably and suffers a drop in clean accuracy. Such turning points occur earlier in our method, making it a trade-off between model accuracy and robustness. We argue that this phenomenon comes from the fact that the clean data distribution can be treated as a marginal distribution in the previous adversarial training, where the model can still fit the original samples. 

\subsection{Time Consumption}\label{sec:time}
In section 2, we analyze the positive correlation between training steps and model performance in adversarial training. Such trade-off in efficiency and effectiveness comes from the complex search process to find the optimal perturbation. DSRM circumvents this issue by providing upper-bound estimates with only clean data. To further reveal the strength of DSRM besides its robustness performance, we compare its GPU training time consumption with other adversarial training methods.
As is demonstrated in Table \ref{tab:3}, the time consumption of DSRM is superior to all the comparison methods. Only TAVAT \cite{li2021token} exhibits similar efficiency to ours (with about 30\% time growth on SST-2 and IMDB). TAVAT neither contains a gradient ascent process on the embedding space, but they still require the construction of additional adversarial data. More experimental details are summarized in Appendix \ref{app:time}.

\begin{table}[thbp]
\centering
\resizebox{0.9\columnwidth}{!}{%
\begin{tabular}{l|c|c|c}
\hline\hline
\textbf{Methods} & \textbf{SST-2} & \textbf{IMDB} & \textbf{AG NEWS} \\ \hline
Finetune & 227 & 371 & 816 \\ \hline
\textbf{DSRM} & \textbf{607} & \textbf{1013} & \textbf{2744} \\ \hline
TAVAT & 829 & 1439 & 2811 \\ \hline
FreeLB & 911 & 1558 & 3151 \\ \hline
PGD & 1142 & 1980 & 4236 \\ \hline
FreeLB++ & 2278 & 3802 & 5348 \\ \hline\hline
\end{tabular}%
}
\caption{GPU time consumption (seconds) of training one epoch on the whole dataset.}
\label{tab:3}
\vspace{-0.7em}
\end{table}
% \paragraph{Training Time Measurement Protocol}
% We measure the training time of each method on GPU and exclude the time for I/O. Each method is run three times and reports the average time. For a fair comparison, every model is trained on a single NVIDIA RTX 3090 GPU with the same batch size for each dataset (8 for IMDB and 32 for the other two datasets).

\subsection{Trade-offs in Standard Adversarial Training}
\begin{figure}[t]
 \centering
 \subfigure[]{
  \begin{minipage}[b]{0.225\textwidth}
   \includegraphics[width=1\textwidth]{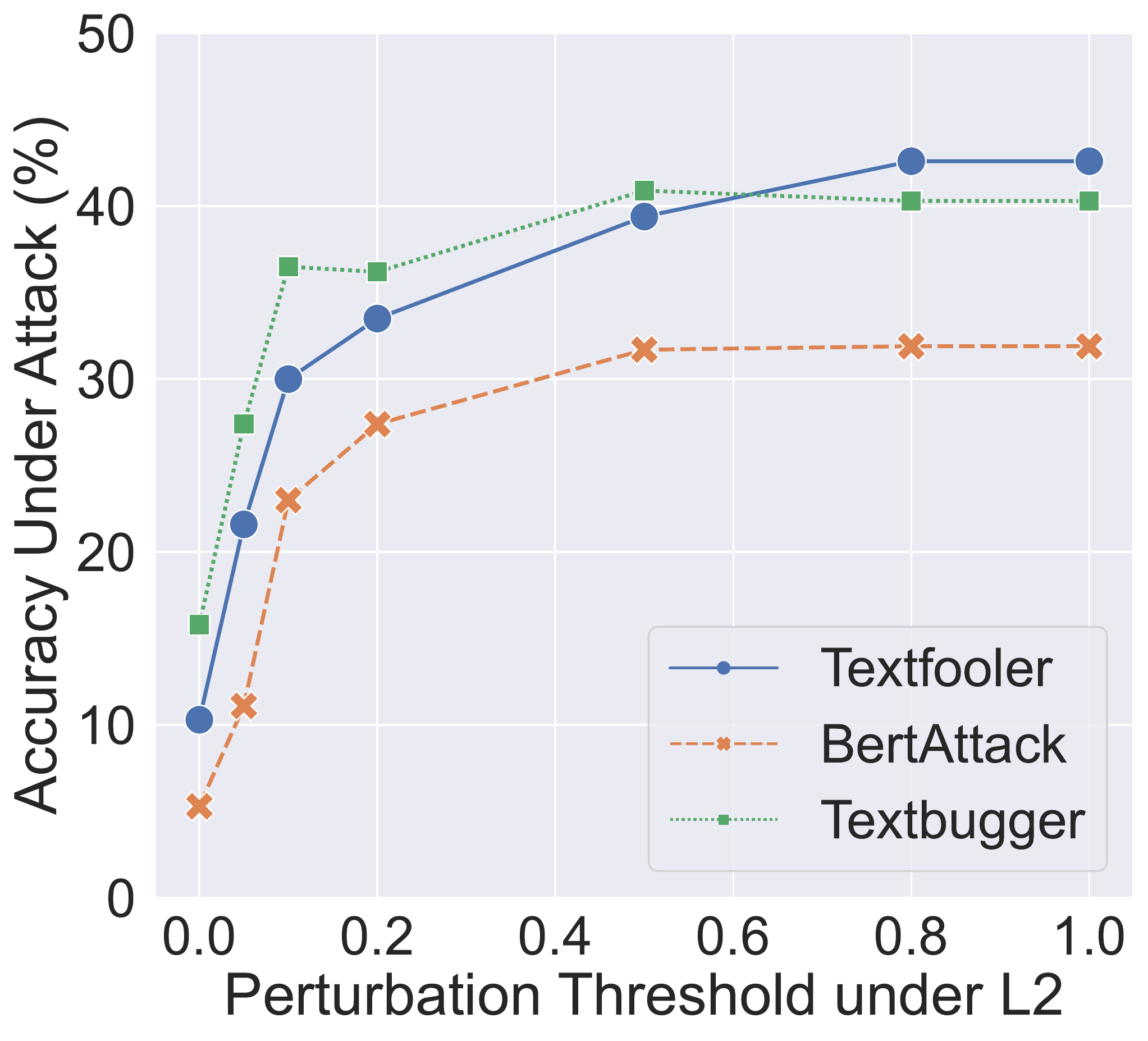}
  \end{minipage}
 }
     \subfigure[]{
      \begin{minipage}[b]{0.226\textwidth}
       \includegraphics[width=1\textwidth]{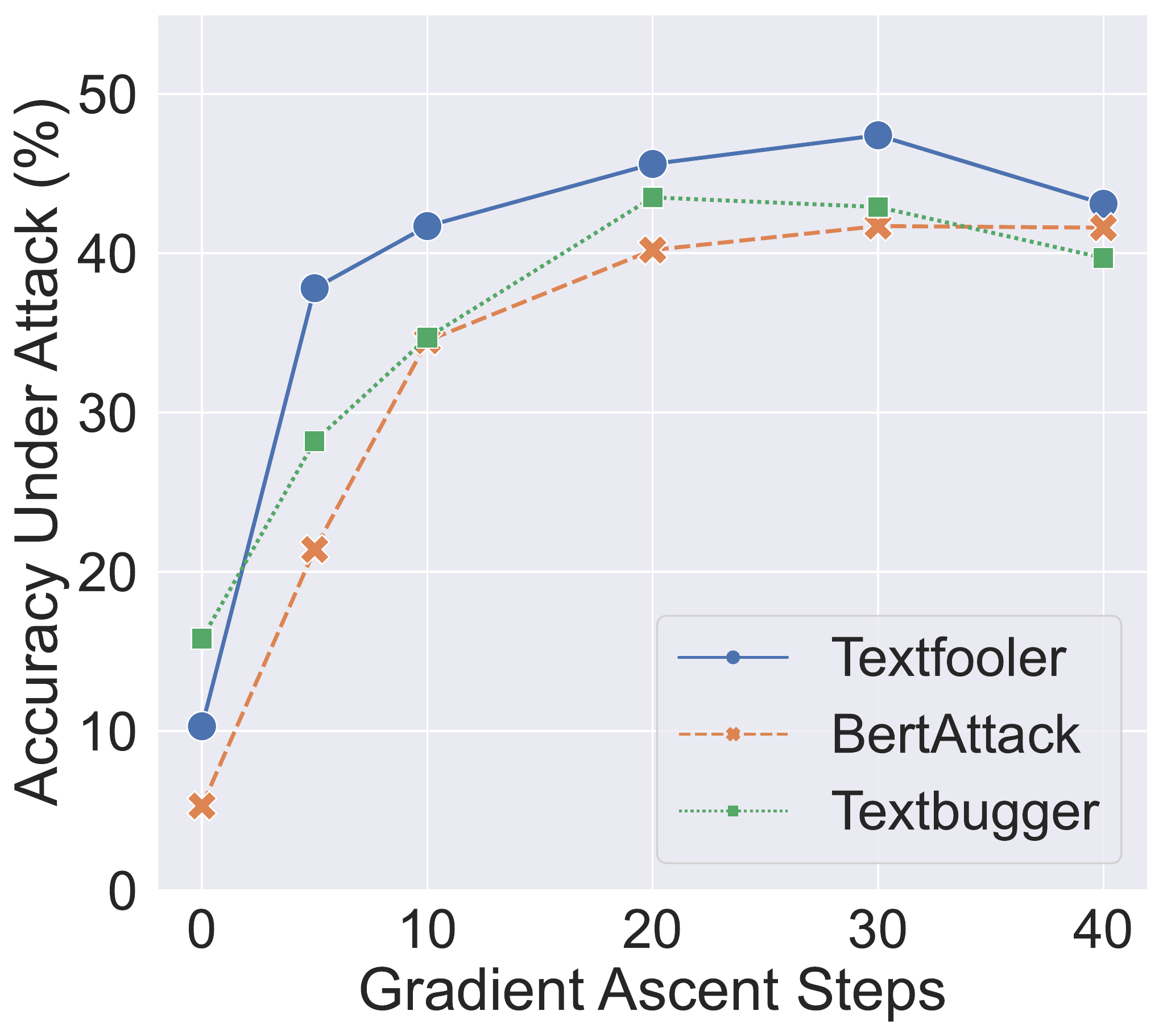}
      \end{minipage}
     }
 \caption{The impact of different values of the perturbation threshold (a) and ascent steps (b) on IMDB dataset. We show the accuracy score of the FreeLB \cite{zhu2019freelb} algorithm under three attack methods. Sub-figure (a) uses a 10-step gradient ascent with different constraints on the $l_{2}$ norm of perturbations. In sub-figure (b), each step introduces a perturbation of length 0.05.}
\label{fig: analysis}
\vspace{-1em}
\end{figure}

In this section, we further discuss the trade-off between computational cost and performance in vanilla adversarial training. We empirically show that larger perturbation radii and steps enhance the effectiveness of textual adversarial training. Similar phenomena are previously found in image datasets by 
\citet{zhang2019theoretically} and \citet{gowal2020uncovering}. The experimental results for these two modifications are shown in Figure \ref{fig: analysis}.

% 1. 更大的步数能提升效果
% 2. 省略对范数的约束能提升效果

In sub-figure (a), relaxing perturbation threshold remarkably increases the model robustness and only suffers a slight decrease when the threshold is larger than 0.6 for Textbugger. In subfigure (b), as the value of steps grows, the models' accuracy under attack increases until they reach their peak points. Subsequently, they begin to decline as the number of steps increases consistently. Notably, the optimal results are 4-10\% higher in (b) relative to (a), demonstrating that a larger number of steps is necessary to achieve optimal robustness.

We give a possible explanation for the above performance. We describe the standard adversarial training as exploring potential adversarial samples in the embedding space. When the step number is small, the adversarial sample space is correspondingly simple, causing the model to underestimate the adversarial risks. A broader search interval can prevent these defects and achieve outstanding robustness as the number of steps grows.

However, these best results occur late in the step growth process. As shown in (b), a defence model needs 30 steps (about ten times the time cost) for Textfooler, 20 for Textbugger, and 40 for BertAttack to achieve optimal performance. This drawback considerably reduces the efficiency and practicality of adversarial training.

\section{Conclusion}
In this paper, we delve into the training objective of adversarial training and verify that the robust optimization loss can be estimated by shifting the distribution of training samples. Based on this discovery, we propose DSRM as an effective and more computationally friendly algorithm to overcome the trade-off between efficiency and effectiveness in adversarial training. DSRM optimizes the upper bound of adversarial loss by perturbing the distribution of training samples, thus circumventing the complex gradient ascent process. DSRM achieves state-of-the-art performances on various NLP tasks against different textual adversarial attacks. This implies that adversarial samples, either generated by gradient ascent or data augmentation, are not necessary for improvement in adversarial robustness. We call for further exploration and understanding of the association between sample distribution shift and adversarial robustness.

\section*{Acknowledgements}
The authors wish to thank the anonymous reviewers for their helpful comments. This work was partially funded by National Natural Science Foundation of China (No.61976056,62076069) and Natural Science Foundation of Shanghai (23ZR1403500).

\section{Limitations}
This section discusses the potential limitations of our work. 
This paper's analysis of model effects mainly focuses on common benchmarks for adversarial defence, which may introduce confounding factors that affect the stability of our framework. Therefore, our model's performance on more tasks, $e.g.$, the MRPC dataset for semantic matching tasks, is worth further exploring. In addition, the present work proposes to conduct adversarial training from the perspective of estimating the overall adversarial loss. We expect a more profound exploration of improving the accuracy and efficiency of such estimation. We are also aware of the necessity to study whether the properties of traditional methods, such as the robust overfitting problem, will also arise in DSRM-based adversarial training. We leave these problems to further work.

\bibliography{anthology,custom}
\bibliographystyle{acl_natbib}

\newpage
\appendix

\section{Dataset Statistics}
\label{app: dataset}
\begin{table}[h]
\begin{tabular}{@{}cccc@{}}
\toprule
\multicolumn{1}{l}{\textbf{Dataset}} & \multicolumn{1}{l}{\textbf{Train/Test}} & \multicolumn{1}{l}{\textbf{Classes}} & \multicolumn{1}{l}{\textbf{\#Words}} \\ \midrule
\textbf{SST-2}                       & 67k/1.8k                                & 2                                    & 19                                   \\
\textbf{IMDB}                        & 25k/25k                                 & 2                                    & 268                                  \\
\textbf{AG NEWS}                     & 120k/7.6k                               & 4                                    & 40                                   \\
\textbf{QNLI}                        & 105k/5.4k                               & 2                                    & 37                                   \\ \bottomrule
\end{tabular}
\caption{Statistics of datasets. In our experiments, we partition an additional 10 per cent of the training set as the validation set to calculate the DSRM of the model}
\end{table}

\section{Experimental details}
\label{app:hyper}
In our experiments, we calculate the sample weights by gradient ascending mean loss to a fixed threshold. The weight of each sample in the normal case is 1/n, where n is the size of a batch.
We fine-tune the BERT-base model by the official default settings. 
For IMDB and AGNews, we use $10\%$ of the data in the training set as the validation set.
The optimal hyperparameter values are specific for different tasks,
but the following values work well in all experiments:
\paragraph{Batch Size and Max Length:} We use batch 16 and max length 128 for SST-2, QNLI, and AG NEWS datasets. For the IMDB dataset, we use batch 8 and max length 256 as its sentence are much longer than other datasets. 
\paragraph{Perturbation Thresholds $\varepsilon$:} $\left[0.8, 1, 1.2, 1.5 \right]$. Weights are truncated when the adversarial loss is greater than the threshold.
\paragraph{Evaluation Settings:}  
For SST-2, we use the official test set, while for IMDB and AGNews, we use the first 1000 samples in the test set to evaluate model robustness.
All three attacks are implemented using TextAttack\footnote{https://github.com/QData/TextAttack} with the default parameter settings.

\section{Training time measurement protocol}\label{app:time}
We measure the training time of each method on GPU and exclude the time for I/O. Each method is run three times and reports the average time. For a fair comparison, every model is trained on a single NVIDIA RTX 3090 GPU with the same batch size for each dataset (8 for IMDB and 32 for the other two datasets).

\end{document}